\documentclass[twocolumn]{article}
\usepackage{graphicx} % Required for inserting images
\usepackage{subcaption}
\usepackage[margin=0.59in]{geometry}
\usepackage{multirow}
\usepackage{hyperref}
\usepackage[toc,page]{appendix}

\usepackage{algpseudocode}
\usepackage{algorithm}

\usepackage[T1]{fontenc}
\usepackage{libertinus}
\usepackage{libertinust1math}
\usepackage{libertine}
\usepackage{titlesec}

\usepackage{url}
\usepackage{cuted}

\usepackage{booktabs}

\titleformat*{\section}{\Large\sffamily}
\titleformat*{\subsection}{\large\sffamily}
\titleformat*{\subsubsection}{\normalsize\sffamily}

\makeatletter
\renewcommand{\maketitle}{%
    \begin{center}
        \includegraphics[width=0.22\textwidth]{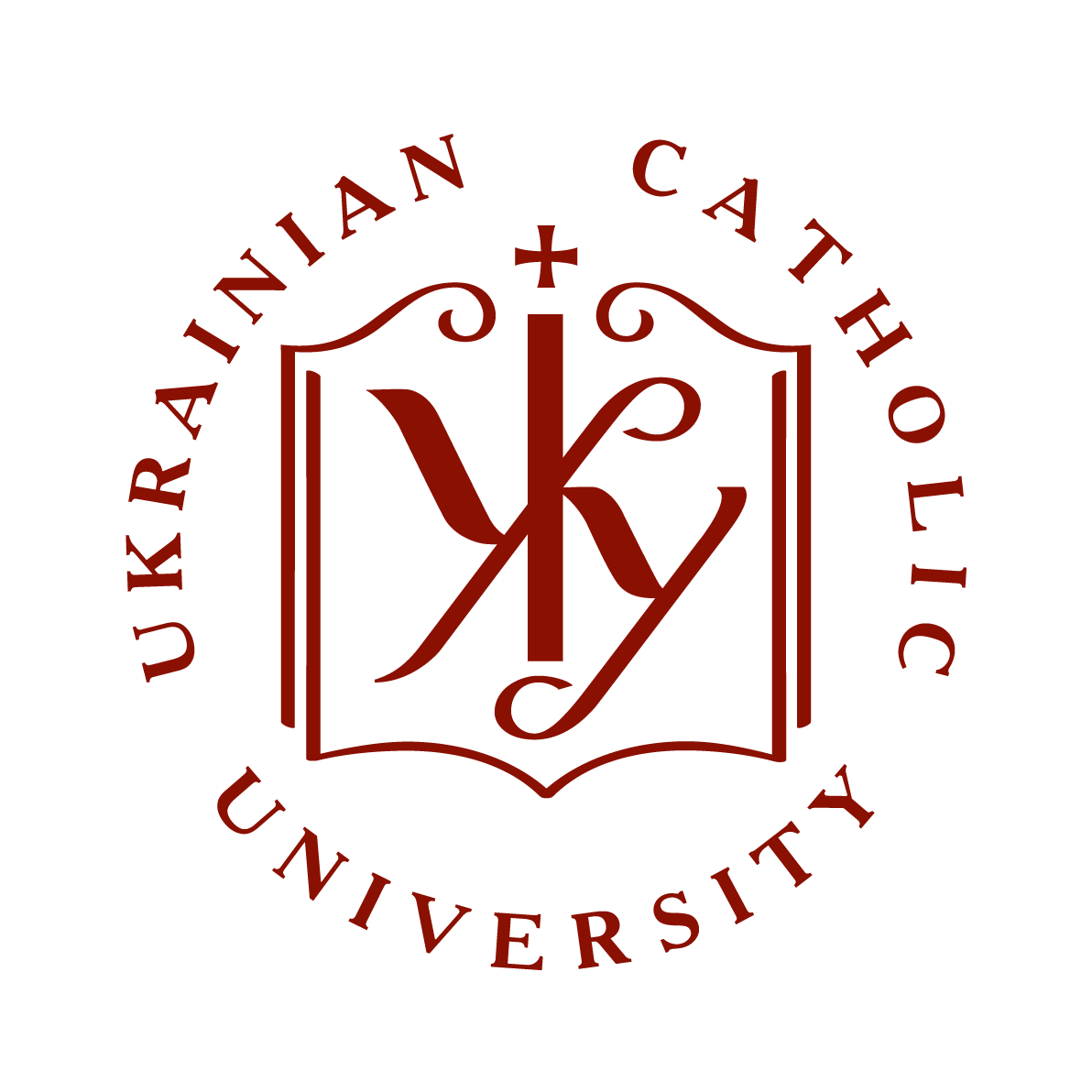}\\[0.8em]
        {\Huge\sffamily \@title}\\[1em]
        {\large\sffamily \@author}\\[0.5em]
    \end{center}
}
\makeatother

\usepackage{blindtext}

\usepackage[round]{natbib}

\title{SLAM in Low-Light Environments: Project Report}
\author{Oleh Basystyi, Anna Stasyshyn, Oleksandr Kosovan, Yaroslav Prytula\\[0.5em]
\small Applied Sciences Faculty, Ukrainian Catholic University\\
\small Lviv, Ukraine\\
\vspace{0.5em}
{\small
Conducted as part of the certification program
``Off-Road Visual Navigation: Development and Evaluation of Systems in
Challenging Environments'' at the Faculty of Applied Sciences,
Ukrainian Catholic University, in collaboration with the UCU UGV Club.}}

\begin{document}

\twocolumn[\maketitle]

\begin{abstract}

% \textit{Simultaneous localization and mapping (SLAM) is one of the fundamental problems in robotics, as it enables autonomous operations in real-world scenarios. 
% However, SLAM's performance is subject to diverse hardware and environmental challenges, such as sensor error, dynamic elements, blurriness, high-speed setups, oscillations, and harsh imaging conditions. 
% These factors can significantly degrade the accuracy of localization and mapping, often leading to accumulated drift, tracking failures, or incomplete maps.}

% \textit{Low-light environments pose a significant challenge for SLAM systems, as visual sensors experience reduced contrast, increased noise, and motion blur. 
% These conditions make the core components of SLAM, namely feature extraction and feature matching, less reliable, thereby degrading map quality and localization accuracy. 
% Alternative sensors, such as LiDAR or depth cameras, can diminish the effect of these issues. However, they increase system cost, power consumption, and integration complexity. 
% For unmanned ground vehicles (UGVs) operating in nighttime terrain or light-restricted zones, robust low-light SLAM is essential for reliable navigation and mapping.}

% \textit{Based on these observations, our aim was to explore whether modern vSLAM systems using standard RGB cameras, without specialized sensors, can achieve stable, sufficiently accurate localization and mapping under low-light conditions.}

% motivation
Simultaneous localization and mapping (SLAM) is one of the fundamental problems in robotics, as it enables autonomous operations in real-world scenarios.
Under low illumination, reduced contrast, sensor noise, and motion blur
degrade both feature extraction and feature matching, while compensating with
LiDAR, depth, or thermal sensors raises cost, power draw, and integration
complexity.
Existing benchmarks remain dominated by well-lit indoor or daylight
sequences, leaving open how far SLAM with standard RGB cameras can be pushed in the
dark.
% results
We benchmark six systems spanning the feature-based, direct, filter-based, and
learning-based paradigms---ORB-SLAM3, DSO, Kimera-VIO, OpenVINS, DPVO, and
DPV-SLAM---on five LaMARia sequences of varying difficulty and illumination,
reporting absolute and relative pose error alongside control-point recall.
Kimera-VIO is the only system to track all five sequences to completion, combining the lowest relative pose error with steadily growing absolute error due to the absence of loop closure; DPVO and DPV-SLAM never lose tracking but incur absolute errors of roughly 100 m under low light; and the classical monocular pipelines (ORB-SLAM3, DSO) together with the filter-based OpenVINS fail outright or diverge on most of the harder and low-light sequences. 
The results suggest that RGB-only SLAM maintains stable low-light tracking only when both inertial fusion and global optimization are present. 
Closing the remaining gap will likely require low-light-specific learned front-ends or a return to complementary sensing.

\end{abstract}

\section{Introduction}

\subsection{SLAM}

Simultaneous localization and mapping~\citep{thrun2005probabilistic} is a method for autonomous navigation of mobile robots, including unmanned ground vehicles (UGVs). 
It defines a state estimation problem in which the autonomous system must determine its location in the environment while generating a map representation of the environment.  

Early SLAM applications relied on filtering algorithms, such as the Kalman filter~\citep{smith1990ekfslam, dissanayake2001solution}, which iteratively updated a joint state vector containing both the landmark poses and the agent's (camera) aggregated pose.
However, as the state vector grows, these methods incur exponentially increasing computational costs and accumulate random walk noise, leading to drift.

Consequently, modern implementations have shifted towards optimization-based (or keyframe-based) systems~\citep{orb_slam, orb_slam_2, orb_slam_3}.
These systems separate the SLAM pipeline into two parallel threads: localization (the front-end), and mapping (the back-end).

\subsection{SLAM in UGV navigation}

To navigate autonomously, a UGV requires continuous awareness of its exact position relative to its environment. 
Even when a mission is initialized with known global coordinates (such as a GPS starting point) and a defined target destination, the robot must reliably determine its current state and location throughout the operation to navigate effectively. 
This continuous state estimation forms the core operational requirement of SLAM.

Beyond localization, the UGV must dynamically interact with its surroundings, introducing the critical need for exploration and path planning. 
While the UGV may have a predetermined target, real-world environments are unpredictable and often contain unknown or moving obstacles that block the initial route. 
To handle this, the map generated by the SLAM pipeline is used by third-party processes, such as path-planning algorithms~\citep{hart1968astar, lavalle1998rrt, macenski2020nav2}, allowing the system to safely explore, recalculate, and adapt its trajectory on the fly.

However, path planning and obstacle avoidance require a generated map. 
Although sparse maps, which consist of scattered 3D map points, are sufficient for tracking the camera's pose and localization, they lack the structural detail needed to detect physical barriers. 
Therefore, for a UGV to successfully avoid obstacles and replan paths, the system must generate dense or semi-dense maps. 
These dense representations provide a comprehensive representation of the scene's geometry, supplying the necessary volume and structural context for the UGV to navigate safely.

\section{Motivation}

The discussion above makes the choice of SLAM as a core technology well-motivated. 
However, it also reveals a critical gap: the environments where UGVs are most needed are precisely those where standard visual SLAM is least reliable.

Industry and dual-use applications routinely demand 24-hour autonomy. 
Emergency response robots must function in power-outage conditions. 
Agricultural robots operate at dawn and dusk. Search-and-rescue systems are deployed in collapsed or smoke-filled structures where artificial lighting is absent or inconsistent. 
In all of these scenarios, the dense and semi-dense maps on which UGV path planning depends must be built from a severely degraded visual signal. 
Despite this practical urgency, most SLAM benchmarks and system evaluations are conducted under well-lit indoor or daylight conditions. 
The gap between the environments in which SLAM is typically evaluated and those in which it must actually operate motivates this work.

Prior benchmarks evaluate SLAM predominantly on well-lit indoor datasets such as TUM RGB-D~\citep{2012bench} and EuRoC MAV~\citep{burri2016euroc}, or outdoor daylight driving sequences such as KITTI~\citep{geiger2012kitti}; none systematically study the combined degradation of ultra-low illumination, dust haze, and motion blur encountered in operational conditions.

Furthermore, adding specialized sensors, such as LiDAR, depth cameras, or thermal imagers, to close this gap increases system cost, power consumption, and integration complexity, which can be prohibitive for lightweight UGV platforms. 
% This raises the central question we aim to address: can modern visual SLAM systems using only standard RGB cameras achieve stable and sufficiently accurate localization and mapping under low-light conditions, and how much of that gap can be closed through algorithmic choices alone.

This motivates the following research questions, which structure the remainder of the report:

\begin{itemize}
    \item \textbf{RQ1}: Can modern visual SLAM systems using only standard RGB cameras
    achieve stable and sufficiently accurate localization and mapping under low-light
    conditions?
    \item \textbf{RQ2}: How does performance degrade across SLAM paradigms --- feature-based, direct, filter-based, and learning-based --- as sequence difficulty and illumination worsen?
    \item \textbf{RQ3}: How much of the low-light gap can be closed through algorithmic
    choices alone, without the addition of specialized sensors?
\end{itemize}

We address \textbf{RQ1} and \textbf{RQ2} empirically, by benchmarking six representative systems on five LaMARia sequences of varying difficulty and illumination (Section~\ref{subsec:results}), and \textbf{RQ3} by relating the observed failure modes to the algorithmic mitigations surveyed in Section~\ref{sec:methods} (Section~\ref{sec:limitations}).

\section{Background}

The SLAM problem can be formalized mathematically. 
Given a sequence of controls $u_{1:t}$ and observations $z_{1:t}$, the goal is to estimate the robot trajectory $x_{1:t}$ and the map $m$. 
Formally, the problem is to compute the posterior~\citep{thrun2005probabilistic}:

$$
P(x_{1:t}, m \mid z_{1:t}, u_{1:t}).
$$ 

Using Bayes' rule, the posterior can be written as:

$$
P(x_{1:t}, m \mid z_{1:t}, u_{1:t}) = \eta P(x_0) \prod_{k=1}^{t} \underbrace{P(x_k \mid x_{k-1}, u_k) P(z_k \mid x_k, m)}_{\text{Motion \& Observation Models}},
$$
where $\eta$ is a normalization constant. This formulation shows that SLAM factorizes into motion constraints between consecutive poses and observation constraints linking poses with the map. The objective is to find the maximum a  posteriori (MAP) estimate:

$$
(x_{1:t}^*, m^*) = \arg\max_{x_{1:t}, m} P(x_{1:t}, m \mid z_{1:t}, u_{1:t}).
$$

Modern SLAM pipelines implement this estimation using two primary abstraction layers: localization (the front-end) and mapping (the back-end).

\subsection{Localization}

Localization, or the front-end, comprises the lower abstraction layer where raw sensor inputs are abstracted into a model used for tracking and incremental localization. 
While the ultimate goal is global consistency, the tracking thread focuses on computing a local trajectory by determining the set of relative rigid-body transformations $\mathbf{T}_{k-1,k} \in SE(3)$ between subsequent frames (we denote elements of $SE(3)$ in bold throughout, with subscripts for time/frame index, superscripts for reference frame). 

Visual localization methods are broadly classified based on their optimization constraints: indirect (feature-based) methods~\citep{orb_slam, orb_slam_2}, and direct (appearance-based) methods~\citep{dso}. 
The former recovers motion ($\mathbf{T}_{k-1,k}$) by minimizing the reprojection error over geometric constraints between matched features, while the latter estimates motion directly from pixel-intensity information by minimizing photometric error.

Additionally, the front-end incorporates loop closure detection and relocalization modules~\citep{7747236}. 
Loop detection is critical for identifying revisited areas to correct accumulated drift, typically involving image representation (e.g., Bag-of-Words)~\citep{orb_slam}, candidate selection, and geometric verification.

\subsection{Mapping}
The back-end is responsible for inferring from the front-end data to generate a consistent map and optimize the overall trajectory. 
The structure and processing of the map generally follow one of two paradigms, which are filter-based approaches and optimization-based approaches.

The filter-based approaches~\citep{thrun2005probabilistic, kalman} are formulated with variants of the Bayes filter. 
These methods marginalize past measurements to maintain a compact joint state vector of landmarks and the current camera pose. 
However, they suffer from exponential computational growth and noise accumulation over time.

The optimization-based methods~\citep{STRASDAT201265} are considered the current de facto standard; these systems solve the "complete graph" of constraints over a subset of frames known as keyframes~\citep{orb_slam, orb_slam_2, orb_slam_3}. 
Keyframes are selected based on criteria such as distance, parallax, or covisibility.

Optimization is typically implemented via factor graphs, utilizing algorithms like Bundle Adjustment (BA)~\citep{bundleadjustment}, which optimizes both camera poses and 3D map points, or Pose Graph Optimization (PGO), which focuses on rigid-body transformations. 
The resulting map can vary in density, ranging from sparse maps consisting of 3D map points to dense maps used for full scene representation. 
Recent advancements are driving a shift toward learning-based pipelines, aiming for end-to-end differentiability in both localization and mapping modules.

To fully understand the landscape of 3D mapping, we survey the variety of scene representations~\citep{deng2026best3dscenerepresentation} in use today, ranging from classic geometric forms to modern neural approaches.

The fundamental classic representation is the point cloud, which is an unordered sparse set of 3D points that is usually derived directly from depth sensors or SLAM tracking. 
While point clouds are highly efficient for simple localization, their discrete nature makes them highly prone to occlusion and a lack of structural volume.

To address the limitations of point clouds' lack of structure, systems often use voxel grids~\citep{octomap}, which divide 3D space into discrete volumetric cells that can store density or occupancy. 
Though highly structured, they entail a computational trade-off between memory consumption and map resolution.

To achieve continuous geometric boundaries without the heavy memory footprint of high-resolution voxels, mapping pipelines frequently convert these discrete representations into meshes. 
Meshes construct continuous surfaces from discrete points, making them suitable for collision checking and physics simulation, though they lack view-dependent appearance data.

Scene graphs~\citep{3dscenegraph} structure the mapped environment as a hierarchical multigraph, fusing raw geometry with high-level semantic and spatial relationships between objects. This makes them suitable for complex task planning and human-robot interaction.

Recently, the mapping paradigm has shifted toward neural representations, starting with Neural Radiance Fields (NeRF)~\citep{nerf}. 
NeRF implicitly encodes 3D scenes into the weights of multi-layer perceptrons, mapping continuous spatial coordinates to color and volume density to provide incredible photorealism.

However, NeRFs require significant computational time to train and render, making real-time robotic deployment difficult. 
3D Gaussian Splatting (3DGS)~\citep{3dgs} has emerged as a powerful alternative. 3DGS explicitly models a scene using millions of learnable 3D anisotropic ellipsoids, achieving high-quality photorealism while enabling extremely fast, real-time rasterization~\citep{matsuki2024gaussiansplattingslam} and map editability.

Latest advancements move beyond explicit geometry and radiance fields by utilizing tokenizer representations or foundation models. 
These models encode complex visual, semantic, and geometric scene elements into a sequence of discrete tokens processed by transformer networks~\citep{wang2024dust3rgeometric3dvision, wang2025vggtvisualgeometrygrounded, dinov3}. 

\section{Methods overview}\label{sec:methods}

\subsection{Keypoint based SLAM}

The reliability of any visual SLAM system depends on its ability to identify and describe \textit{features}, a term often used interchangeably with key points. 
Features are distinctive parts of an image, such as corners, edges, or textures. 
This process is divided into two distinct steps: detection and description. 
A detector identifies interest points (i.e., keypoints), which are preferably robust to rotation, scale, and the effects of illumination on their color appearance and sharpness. 
Once a point is identified, a descriptor converts the local neighborhood  of that feature into a numeric vector, which must be invariant to environmental shifts to allow for consistent recognition across multiple views. Once features are successfully described, the system must establish correspondences between frames, which is the process known as data association.

\subsubsection{Feature extractors and descriptors}

Feature extraction serves as the critical bridge between raw sensory input and geometric motion estimation by identifying points of interest that are consistent across varying perspectives. 
Traditional approaches to this problem, such as FAST~\citep{mlcorner} (Features from Accelerated Segment Test) and ORB~\citep{orb} (Oriented FAST and Rotated BRIEF~\citep{brief}), rely on high-speed heuristic tests of local pixel intensities. 
FAST identifies corners by evaluating a circle of sixteen pixels around a candidate point, designating it a feature only if a contiguous arc of pixels is significantly brighter or darker than the center. 
While computationally efficient, FAST is not by itself rotation invariant. 
ORB addresses this limitation by calculating the intensity centroid of the identified patch to assign a principal orientation, subsequently using BRIEF to generate binary descriptors that remain robust during camera rotation. 

Other classical algorithms, such as SIFT~\citep{sift} (Scale Invariant Feature Transform) and SURF~\citep{10.1007/11744023_32} (Speeded Up Robust Features), offer better reliability due to scale and luminance invariance. 
SIFT achieves this by detecting features across a continuous scale space and assigning orientations based on local image gradients. 
However, the high-dimensional floating-point descriptors and multiple processing stages required by SIFT and SURF result in significant computational overhead. This makes them generally unsuitable for real-time robotic tasks, such as UGV navigation. 
While they produce higher-quality matches than ORB in some scenarios, the trade-off in execution speed remains a bottleneck for practical SLAM implementations.

The current state of the art has shifted toward deep learning-based extractors, most notably SuperPoint~\citep{superpoint}, to overcome the specific failures of geometry-based methods in more complex or ambiguous scenarios. 
SuperPoint utilizes a self-supervised interest point detector and descriptor network trained to recognize the same landmarks under extreme shifts in perspective, illumination, and color appearance. 

\subsubsection{Feature matchers}

Once features are extracted, the SLAM front-end must establish accurate correspondences between descriptors in successive frames to estimate the relative rigid body transformation. 
Traditional approaches typically rely on brute-force matching, which searches for the best match by comparing a feature in the first image against all features in the second based on distance metrics. 
It calculates Hamming distance for binary descriptors like ORB, and the L2 distance for floating-point descriptors like SIFT. 

To maintain real-time performance as the number of landmarks increases, the SLAM pipeline often employs Approximate Nearest Neighbor~\citep{10.1145/293347.293348} (ANN) algorithms. 
These methods trade a negligible amount of precision for a significant increase in search speed by utilizing spatial indexing structures, most notably KD-trees. 
A KD-tree partitions the high-dimensional descriptor space into a hierarchical tree, allowing the matcher to quickly prune large portions of the search space that cannot contain a close match. 
While ANN methods are highly efficient, especially with accelerated implementations like FANN~\citep{fast}, they remain prone to errors caused by "lookalikes", which occur when distinct physical points have nearly identical descriptors, leading to outliers that must be filtered through subsequent geometric verification steps, such as RANSAC.

To move beyond the limitations of isolated descriptor comparisons, modern pipelines are increasingly incorporating deep learning matchers. 
Unlike classical algorithms that treat each keypoint as an independent entity, DL approaches like LightGlue~\citep{sarlin2023lightglue} utilize graph neural networks and self-attention mechanisms to learn the spatial and geometric relationships between points. 
By analyzing the global layout of the scene, these matchers can distinguish between visually similar points based on their relative positions, significantly reducing the mismatch rate in repetitive or complex environments.

The combination of deep learning feature extractors and matchers is present in a large portion of the currently used approaches, namely the SuperPoint and LightGlue combination. 
However, these approaches raise the problem of computational overhead and increased hardware requirements.

\subsection{Low-light localization and mapping}
As discussed above, low-light environments present a challenging task for modern SLAM systems because they heavily rely on visual sensors. Several categories of mitigation have been proposed.
One of the naive approaches involves image preprocessing with CLAHE~\citep{clahe}, which enhances local contrast. However, performance degrades in moderately low-light conditions. 
Newer approaches focus on deep learning preprocessing~\citep{ldfeslam, lfpd} to get a more robust, brightened image.
Another deep learning paradigm considers the usage of feature extractors tailored for low-light environments~\citep{deeplsd}.  

\subsubsection{Effects of low-light}

In visual SLAM and autonomous navigation, the transition from standard indoor lighting (300–500 lux) to low-light environments (10–50 lux) or extreme darkness (below 1 lux) introduces several degradations that fundamentally break the many assumptions about the environment that traditional algorithms rely on. 
Reduced illumination has distinct and measurable effects on the image quality.

As illumination drops, the sensor's electronic noise becomes more prominent relative to the captured light, resulting in "grainy" images that obscure fine geometric details. 
This effect is measured by the signal-to-noise ratio (SNR), defined as the ratio of signal power to noise power.

Reduced contrast leads to the disappearance of sharp edges and corners, which are the primary "landmarks" used by classical feature extractors like ORB or FAST.

To compensate for low light, cameras often increase exposure times; for a moving agent, this results in significant blurring that smears features across multiple pixels. 
In extreme low light (below 1 lux), digital sensors struggle to perceive color accurately, leading to shifted or desaturated palettes that fail color-consistency checks.

For visual SLAM systems, these degradations lead to a high mismatch rate, where the system either fails to extract enough points or incorrectly pairs noisy pixels in consecutive frames, ultimately resulting in localization drift or total tracking failure.

\subsubsection{Preprocessing and image enhancement}

The natural idea to tackle the problem of insufficient illumination is image preprocessing. Several methods are used to fix under-exposure. 

One such approach is EnlightenGAN~\citep{DBLP:journals/corr/abs-1906-06972}, which is trained with unpaired supervision.
This allows the model to learn how to brighten and denoise images without requiring a dataset of perfectly aligned dark and bright training pairs, which are often difficult to capture in real-world scenarios.

The Bread~\citep{10.1007/s11263-022-01667-9} architecture is designed to handle more severe degradations by decomposing the enhancement process into three distinct sub-tasks: noise suppression, illumination adjustment, and color restoration. 
It operates by converting the input into a luminance-chrominance space, allowing the system to suppress noise in the brightened luminance channel independently.

Zero-DCE~\citep{DBLP:journals/corr/abs-2001-06826} formulates image enhancement as the estimation of pixel-wise, high-order nonlinear curves. 
Unlike many other deep learning approaches, it trains entirely without reference images—either paired or unpaired—by relying on a set of non-reference loss functions.

Dual Illumination Estimation~\citep{DBLP:journals/corr/abs-1910-13688} addresses robust exposure correction by simultaneously tackling under- and over-exposure. 
It predicts a forward illumination map for the original image to correct dark regions and a reverse illumination map for an inverted version of the input to handle over-exposed areas.

\subsubsection{Robust matchers and extractors}

While image enhancement modules can recover lost illumination, the preprocessed images often retain artifacts such as synthetic noise, color distortion, or smoothed textures. 
These residual degradations cause classical geometric extractors like ORB to produce clustered, unreliable detections, or fail entirely. 
Consequently, a robust low-light SLAM pipeline must integrate perception components that can resist these artifacts.

To address this, the combination of deep-learned keypoints and graph-based matching, most notably the SuperPoint~\citep{superpoint} and SuperGlue~\citep{sarlin20superglue} (or its more efficient successor, LightGlue~\citep{sarlin2023lightglue}) architecture, is widely adopted. 
In low-light scenarios, local pixel gradients are often too weak or noisy to define a distinct corner. 
SuperPoint overcomes this with learned, higher-level structural context. 
Rather than relying on rigid heuristic tests, it is trained with homographic adaptation to recognize the underlying geometry of a scene, allowing it to reliably detect repeatable landmarks even in visually ambiguous or under-exposed regions.

Once these features are extracted, establishing accurate correspondences in the dark becomes the next challenge. 
In low-light environments, distinct physical points frequently produce nearly identical, low-contrast local patches. 
Traditional nearest-neighbor matchers fail because they evaluate these descriptors in isolation. 
The SuperGlue and LightGlue architectures solve this by utilizing self- and cross-attention mechanisms to evaluate the global geometric layout of the scene. 
By contextualizing a keypoint within the broader structure of the image, these matchers can accurately resolve ambiguities and reject the "lookalike" outliers that typically dominate dark scenes.

Beyond the SuperPoint and LightGlue paradigm, several alternatives have been explored specifically for light-denied environments. Dense matchers, such as LoFTR~\citep{sun2021loftr} (Local Feature Matching with Transformers), bypass the keypoint detection phase entirely. 
Instead of extracting discrete features, LoFTR establishes coarse pixel-wise matches at a lower resolution and refines them at a higher resolution. This correlation-based approach is exceptionally effective in low-light and texture-less environments where distinct keypoints simply cannot be found. 
However, dense matchers introduce significant computational overhead (which is a more significant problem considering the use of previous stages of battling low-light), making them challenging to deploy for real-time UGV navigation compared to the adaptive, early-exit architecture of LightGlue. 

Other learned extractors like D2-Net~\citep{Dusmanu2019CVPR}, which tightly couples feature detection and description, or ALIKE~\citep{Zhao2022ALIKE} and ALIKED 
~\citep{Zhao2023ALIKED}, offer varying trade-offs between feature density and localization accuracy. 
Nonetheless, for systems requiring a balance between robustness against illumination changes and real-time inference speeds, the SuperPoint and LightGlue pairing currently serves as the optimal baseline for low-light visual SLAM.

\subsubsection{Low-light in other 3D tasks}

One of the latest trends in SLAM is the usage of alternative scene representations during the mapping phase. 
Some of these approaches may help with the low-light task. Specifically, 3DGS, NeRFs, and foundation model approaches have shown significantly better performance for this task.  

\citet{llgaussian} consider the novel-view synthesis problem for low-light scenes. 
As in SLAM, these algorithms fail in low-light conditions because they assume high-quality, well-exposed inputs.

Instead of relying on SfM, the algorithm uses dense priors from a learning-based Multi-View Stereo approach to generate initial point clouds. 
It then applies stochastic pruning and depth-guided refinement to remove artifacts. The authors also reformulate the splatting optimization problem. 
They introduce Dual-branch Gaussian Decomposition. The scene is disentangled into two sets of Gaussians -- intrinsic Gaussians, which capture static reflectance and illumination, and transient Gaussians, which specifically model unstable content like sensor noise and illumination artifacts.  

LL-Gaussian significantly outperforms SoTA methods in low-light sRGB reconstruction. 
It achieves up to 2,000x faster inference and reduces training time to just 2\% compared to NeRF-based methods. 
It maintains fine details and accurate colors even in moonlit environments (0.1 lux) where other methods produce blurry textures or fail to converge. 
Also, the LLGIM initialization achieves accuracy comparable to using ground-truth normal-light data.

Even though this paper does not address SLAM directly, it shows that learning-based MVS can replace traditional feature matching in the dark. 
In addition, the dual-branch decomposition technique may be applicable to GS-based SLAM approaches, potentially improving mapping in the dark while also partially addressing dynamic environments.

\subsubsection{Sensor fusion for light-denied environments}

While deep learning-based image enhancement and robust feature matching significantly extend the operational envelope of visual SLAM, absolute light deprivation or sudden, drastic illumination shifts can still cause catastrophic tracking failures. 
In environments where the visual signal is close to non-existent, algorithmic mitigation alone is insufficient. 
To ensure continuous and safe autonomous UGV operation, the SLAM pipeline must transition from a purely visual approach to a multi-modal sensor fusion architecture, integrating sensors that possess complementary failure domains.

The most fundamental integration is Visual-Inertial SLAM (VI-SLAM), which tightly couples camera data with an Inertial Measurement Unit (IMU). 
The IMU provides high-frequency kinematic measurements—specifically linear acceleration and angular velocity—that are entirely independent of ambient lighting. 
In a tightly-coupled optimization framework, the IMU maintains the short-term ego-motion estimation during transient periods of severe motion blur, sudden glare, or complete darkness. 
While IMU measurements inherently accumulate drift over time through double integration, they successfully bridge the temporal gaps in visual tracking until the camera recovers enough reliable features to correct the accumulated positional error.

For sustained operation in light-denied settings, passive visual sensors are frequently fused with active ranging devices such as LiDAR. 
Because LiDAR emits its own infrared laser pulses to measure distance, it is completely invariant to external illumination, providing highly accurate 3D geometry in pitch-black conditions. 
Fusing these modalities creates a robust mutual-compensation mechanism. 
When visual tracking fails due to darkness, the system seamlessly falls back on LiDAR geometry; conversely, when the UGV navigates geometrically degenerate environments, such as long, featureless tunnels where LiDAR struggles to estimate forward motion, enhanced visual features provide the necessary longitudinal constraints.

Furthermore, emerging sensor technologies like event cameras and thermal imaging are becoming highly relevant for nighttime SLAM. 
Event cameras are asynchronous neuromorphic sensors that record pixel-level brightness changes rather than absolute intensity frames. 
They possess an exceptionally high dynamic range and microsecond latency, rendering them virtually immune to motion blur and highly effective in both extreme darkness and high-dynamic-range scenes. 
Similarly, Long-Wave Infrared (LWIR) or thermal cameras capture heat signatures independent of visible light. 
Incorporating thermal data into the SLAM pipeline not only provides a reliable tracking signal in the dark but also natively assists in identifying and filtering dynamic objects, such as pedestrians or running vehicles, further supporting the static-scene assumptions required for stable mapping.

\subsection{Handling of non-static environments}

As discussed above, the SLAM pipeline works by tracking the same features between consecutive keyframes (for example, pixels corresponding to the same points in 3D space). 
Given such correspondences, it is straightforward to estimate the relative motion $\mathbf{T}_{i,i+1}$. 
However, if the scene is non-static (e.g.\ other moving vehicles, or intense branch wiggling), this estimate may be highly inaccurate. 
This motivates a mechanism to detect and reject invalid correspondences, or to minimize the effect of minor violations during optimization.

\citet{Soares2025var} propose a visual SLAM framework that improves robustness in dynamic scenes by combining semantic correspondence filtering with adaptive robust optimization. 
The method extends the ORB-SLAM3 pipeline. 
To address known dynamic objects, the system employs a lightweight deep-learning object detector (YOLOv4).
The detector identifies bounding boxes corresponding to dynamic classes (e.g., people), and correspondences within these regions are filtered. Depth consistency is additionally used to distinguish between foreground object points and background features, allowing the removal of correspondences associated with moving objects while preserving useful background structure.

The remaining (possibly non-static) observations are optimized using a robust bundle adjustment formulation.
Let $\theta$ denote the set of camera poses and 3D map points, and let $e_i(\theta)$ represent the magnitude of the reprojection residual. 
The SLAM estimation problem is formulated as

$$
\theta^* = \arg\min_{\theta} \sum_{i} \rho(e_i(\theta)).
$$

Instead of using a fixed robust kernel such as the Huber loss, the method employs Barron’s generalized robust loss, defined as

$$
\rho(e;\alpha,c)
=
\frac{|\alpha-2|}{\alpha}
\left[
\left(
\frac{e^2}{|\alpha-2|c^2}+1
\right)^{\alpha/2}
-1
\right],
$$

where $\alpha$ controls the shape of the loss function and $c$ is a scale parameter. 
Different values of $\alpha$ correspond to different classical estimators (e.g., $\alpha=2$ recovers $L_2$, $\alpha=1$ recovers smooth $L_1$), enabling the optimizer to adapt its robustness to the residual distribution, which in turn means that it can cope with different degrees of non-static assumption violation. 
The parameter $\alpha$ is estimated online by minimizing the negative log-likelihood of the residuals:

$$
\alpha^* =
% \arg\min_{\alpha}
% \sum_{i}
% \left[
% \log \tilde{Z}(\alpha)
% +
% \rho(e_i;\alpha,c)
% \right].
% \]
\arg\min_{\alpha}
\sum_{i}
\left[
\log \tilde{Z}(\alpha)
+
\rho(e_i;\alpha,c)
\right],
$$

where $\tilde{Z}(\alpha)$ is a function ensuring normalization of output values.
The optimization alternates between updating $\alpha$ and solving for the SLAM variables using iteratively reweighted least squares. 
Let $K$ denote the set of keyframes in the local window, $P_j$ the set of map points observed in keyframe $j$, $u_{ij}$ the observed pixel location of point $i$ in keyframe $j$, $\pi(\cdot)$ the camera projection function, $X_i$ the 3D position of map point $i$, and $\mathbf{T}_j^{cw}$ the camera-to-world pose of keyframe $j$. The adaptive loss is then applied as:

$$
\{\mathbf{T}_j^{cw}, X_i, \alpha\}^*=
\arg\min_{\{\mathbf{T}_j^{cw}\},\{X_i\},\alpha}
\sum_{j \in K} \sum_{i \in P_j}
\rho\!\left(
\|u_{ij}-\pi(\mathbf{T}_j^{cw}X_i)\|^2; \alpha, c
\right).
$$

By combining deep-learning-based dynamic object detection with adaptive robust optimization, the proposed approach could minimize the effect of dynamic features, allowing the SLAM system to maintain accurate localization and mapping in dynamic environments while preserving real-time performance.

\section{Experiments}

\subsection{Datasets}

Among visual-inertial SLAM datasets, EuRoC~\citep{burri2016euroc} is the classic benchmark. 
Besides IMU and stereo data, it also provides accurate ground truth. However, there are reasons why it is not a suitable dataset choice for our context.  

Firstly, it consists of indoor sequences. This does not reflect our target scenario, which mostly includes military applications.
Secondly, the problem of low-light is not captured in the official dataset. 
Some open-source contributions extend the sequences with variants that artificially introduce varying levels of visual degradation. 
Having the same high-quality ground truth for each sequence variant makes it possible to isolate the effects of illumination.
However, these synthetic degradations are often not physically realistic, and the footage does not account for the camera's exposure adaptation.

Other options include the TUM VI benchmark~\citep{Schubert_2018}, which includes indoor and outdoor sequences but focuses mainly on challenging VIO with high-FOV cameras, and the KITTI benchmark~\citep{geiger2012kitti}, which specializes in driving and provides LiDAR and GPS but whose setting differs substantially from the operational environments considered in this work.

The dataset we have chosen for our experiments is the LaMARia~\citep{krishnan2025benchmarkingegocentricvisualinertialslam}. 
It was specifically designed to benchmark visual-inertial SLAM under realistic egocentric operating conditions and includes a wider range of environmental and illumination variations that better reflect our target deployment scenario.

\subsection{Evaluation metrics}

Performance is evaluated using standard visual-inertial SLAM metrics. 
We report the absolute trajectory error (ATE) and relative pose error (RPE).

The absolute trajectory error measures the global consistency of the estimated trajectory. 
Let $\mathbf{T}_i \in SE(3)$ denote the ground-truth pose at time $i$ and $\hat{\mathbf{T}}_i \in SE(3)$ the corresponding estimated pose after trajectory alignment. 
Then, the translational ATE is defined as:

$$
\sqrt{
\frac{1}{N}
\sum_{i=1}^{N}
\left|
\mathrm{trans}
\left(
\mathbf{T}_i^{-1}
\hat{\mathbf{T}}_i
\right)
\right|^2
},
$$

where $\mathrm{trans}(\cdot)$ extracts the translational component of a rigid-body transformation and $N$ is the number of evaluated poses. 
A lower ATE indicates better global localization accuracy and reduced drift accumulation.

To evaluate local motion estimation accuracy, we report the relative pose error. 
For a fixed time interval $\Delta$, the relative pose error $E$ at time $i$ is given by

$$
\mathbf{E}_i=\left(
\mathbf{T}_i^{-1}
\mathbf{T}_{i+\Delta}
\right)^{-1}
\left(
\hat{\mathbf{T}}_i^{-1}
\hat{\mathbf{T}}_{i+\Delta}
\right)
$$

The translational RPE is then computed from $\mathbf{E}_i$ as:

$$
\sqrt{
\frac{1}{M}
\sum_{i=1}^{M}
\left|
\mathrm{trans}(\mathbf{E}_i)
\right|^2
},
$$

where $M$ denotes the number of valid pose pairs separated by $\Delta$. 
Unlike ATE, which measures accumulated drift over the entire trajectory, RPE focuses on short-term motion estimation accuracy and is therefore less sensitive to global alignment errors.

In order to assess both local and global accuracy of the estimated trajectory, we use two complementary metrics introduced in \citep{krishnan2025benchmarkingegocentricvisualinertialslam}. We discuss and report them in the appendix.

\subsection{Methods evaluated}

The experimental evaluation includes the following representative visual(-inertial) SLAM and odometry systems that collectively cover feature-based, direct, and optimization-based approaches.

ORB-SLAM3~\citep{orb_slam_3} is a feature-based visual-inertial SLAM system that extends previous ORB-SLAM versions~\citep{orb_slam, orb_slam_2} with tightly coupled inertial integration. 
It supports monocular, stereo, and RGB-D configurations and incorporates loop closure and map reuse. Due to its strong performance across a wide range of benchmarks, ORB-SLAM3 serves as a widely adopted baseline for visual-inertial SLAM.

Kimera-VIO~\citep{kimera} is the visual-inertial odometry front-end of the modular Kimera framework, which in its full configuration also supports loop closure and metric-semantic mapping. 
We evaluate the VIO-only configuration; its tightly coupled optimization-based backend makes it representative of modern graph-based visual-inertial odometry systems.

Direct Sparse Odometry~\citep{dso} is a direct visual odometry method that estimates camera motion by minimizing photometric error over selected image pixels rather than relying on feature correspondences. 
As direct methods utilize image intensity information directly, they provide an interesting comparison under challenging illumination conditions.

OpenVINS~\citep{openvins} is a visual-inertial state estimation framework based on the Multi-State Constraint Kalman Filter (MSCKF). 
Unlike optimization-based approaches, OpenVINS employs a filtering-based architecture, making it computationally efficient while providing competitive visual-inertial odometry performance.

DPVO~\citep{teed2023dpvo} (Deep Patch Visual Odometry) is a learning-based monocular visual odometry system that represents the scene through a sparse set of image patches rather than individual keypoints or dense photometric residuals. It tracks these patches across frames using a recurrent network that iteratively predicts patch correspondences and feeds them into a differentiable bundle adjustment layer, jointly refining camera poses and patch depths. Because the patch representation and the learned update operator are trained end-to-end, DPVO does not rely on hand-crafted feature detectors, brightness-constancy assumptions, or inertial priors, which makes it an interesting candidate for degraded-illumination sequences where classical front-ends break down. However, as a pure odometry method it maintains only a local optimization window and therefore accumulates drift without any global correction.

DPV-SLAM~\citep{lipson2024dpvslam} extends DPVO into a full SLAM system by adding a backend that performs loop closure and global optimization on top of the DPVO frontend. It couples a proximity-based loop-closure mechanism with a classical pose-graph loop closure, allowing accumulated drift to be corrected when previously visited regions are revisited, while retaining the real-time patch-based tracking of DPVO. This makes DPV-SLAM directly comparable to DPVO as an ablation that isolates the contribution of the global backend, while both remain monocular and inertial-free.

\subsection{Experiment setup}

Experiments were conducted on a subset of the LaMARia~\citep{krishnan2025benchmarkingegocentricvisualinertialslam} dataset consisting of the sequences
\verb|sequence_4_10|,
\verb|sequence_4_11|,
\verb|R_03_easy|,
\verb|R_06_medium|, 
and
\verb|R_09_hard|. 
The selection of a subset of the captures was necessary as evaluating all sequences in the benchmark would be computationally expensive.

The selected sequences include both relatively straightforward and challenging trajectories. 
This allows the evaluation to capture method behavior across various conditions, ranging from scenarios with stable tracking and favorable visual observations to sequences containing more challenging motion patterns, environmental complexity, and visual degradation.

The sequences \verb|R_03_easy|,
\verb|R_06_medium|, 
and
\verb|R_09_hard| 
were selected to capture the varying difficulty levels represented in the dataset. 
This allows us to evaluate whether the performance of the considered methods degrades with track complexity and tracking difficulty. 
The additional sequences were selected because they exhibit substantial illumination variation, which is particularly relevant to the objectives of this work. 
Together, this produces a balanced benchmark without the expense of evaluating all of the scenes.

\subsection{Results}\label{subsec:results}
\begin{table*}[htbp] 
\centering
\begin{tabular}{lccccc}
\toprule
\textbf{Method} & \textbf{easy} & \textbf{medium} & \textbf{hard} & \textbf{low-light 1} & \textbf{low-light 2} \\
 & (\texttt{R\_03\_easy}) & (\texttt{R\_06\_medium}) & (\texttt{R\_09\_hard}) & (\texttt{sequence\_4\_10}) & (\texttt{sequence\_4\_11}) \\
\midrule
DSO             & 0.400 & 11.310 & 17.478$^\dagger$ & \underline{15.473}$^\dagger$ & \textbf{7.147}$^\dagger$ \\
ORB-SLAM3 (mono)& 2.768 & 22.809$^\dagger$ & \textbf{5.099}$^\dagger$ & \textbf{2.185}$^\dagger$ & \underline{8.378}$^\dagger$ \\
Kimera-VIO      & 0.410 & \underline{2.795} & \underline{6.055} & 44.962 & 43.042 \\
OpenVINS        & \textbf{0.259} & 18.997 & fail & fail & 11.163 \\
DPVO            & \underline{0.378} & \textbf{2.123} & 26.267 & 104.874 & 99.453 \\
DPV-SLAM        & 0.483 & 3.637 & 18.460 & 101.665 & 97.482 \\
\bottomrule
\end{tabular}
\caption{Comparison of different VIO/SLAM methods across various sequences. Metric evaluated is RMSE ATE~$\downarrow$ (m).}
\label{tab:method_comparison}
\end{table*}

We report ATE RMSE (m) on the specified sequences. 
A dagger ($\dagger$) marks a partial-coverage track ($<\!50\%$ of GT poses). 
The system lost tracking and the ATE is over the short tracked span only, so it is not comparable to a full-coverage ATE. 
The only sequence fully covered by all approaches is \verb|R_03_easy|.

\begin{table*}[htbp]
\centering
\begin{tabular}{lccccc}
\toprule
\textbf{Method} & \textbf{easy} & \textbf{medium} & \textbf{hard} & \textbf{low-light 1} & \textbf{low-light 2} \\
 & (\texttt{R\_03\_easy}) & (\texttt{R\_06\_medium}) & (\texttt{R\_09\_hard}) & (\texttt{sequence\_4\_10}) & (\texttt{sequence\_4\_11}) \\
\midrule
DSO             & \underline{0.065} & 0.132 & fail & \underline{0.640}$^\dagger$ & fail \\
ORB-SLAM3 (mono)& 0.161 & fail & fail & fail & fail \\
Kimera-VIO      & 0.111 & 0.140 & 0.113 & \textbf{0.370} & \underline{0.224} \\
OpenVINS        & \textbf{0.004} & \textbf{0.032} & fail & fail & \textbf{0.098} \\
DPVO            & \underline{0.065} & \underline{0.082} & \textbf{0.067} & 0.814 & 1.116 \\
DPV-SLAM        & 0.066 & \underline{0.082} & \underline{0.073} & 1.004 & 0.973 \\
\bottomrule
\end{tabular}
\caption{Comparison of different VIO/SLAM methods across various sequences. Metric evaluated is RPE RMSE~$\downarrow$ (m), computed over the same tracked spans as Table~\ref{tab:method_comparison}. $\dagger$ and \emph{fail} follow the same convention as Table~\ref{tab:method_comparison}.}
\label{tab:method_comparison_rpe}
\end{table*}

Table~\ref{tab:method_comparison_rpe} reports the corresponding RPE RMSE for each method and sequence.

As the strongest contestant, Kimera-VIO provides full coverage on all five tracks with competitive ATE. Notably, ATE grows significantly with trajectory length while RPE remains low, a direct consequence of the lack of loop closure to cancel accumulated error. 
As a VIO rather than a full SLAM pipeline, Kimera-VIO is nonetheless a strong choice for this task.

The largest failure case is OpenVINS on \texttt{sequence\_4\_10} and \texttt{R\_09\_hard}.
While some approaches lost tracking over time, and still had a well-tracked prefix (an interval from the start of the sequence up to the point of significant failure), OpenVINS did not produce such an initial interval at all. 
Before the failure, the ATE already exceeded the values demonstrated by other approaches significantly. 
This is consistent with filter divergence, which in this dataset can be triggered by handheld motion together with feature loss and outliers arising from occlusions and complex, moving platforms.

The DPVO and DPV-SLAM offer great robustness, with no tracking failure despite not relying on inertial priors. 
However, this comes at a tradeoff of reduced accuracy compared to the successful VIO counterpart, Kimera-VIO.

DSO struggles with the task because direct methods assume brightness constancy, while the dataset includes handheld motion blur, illumination change, and low texture. 

ORB-SLAM3 does not perform well either, since its feature-based front-end fails under low texture, blur, or pure rotation, where sufficient parallax is required. 
After losing tracking, a new map is initialized at a fresh, arbitrary scale, making recovery of the trajectory fragments unstable.

\section{Limitations}\label{sec:limitations}

The evaluated systems span feature-based, direct, filter-based, and learning-based paradigms, and each carries limitations that are rooted in its core assumptions. These limitations are not incidental to our benchmark; they surface directly in the reported ATE and RPE (Tables~\ref{tab:method_comparison} and~\ref{tab:method_comparison_rpe}) and control-point recall (Table~\ref{tab:lamaria-cp}), and they shape how each method could be adapted to the low-light UGV domain.

ORB-SLAM3 inherits the fundamental limitation of indirect front-ends: it requires a sufficient number of repeatable, well-localized keypoints and adequate parallax between views. Under low illumination, motion blur, low texture, and near-pure rotation, the ORB detector produces too few stable corners, tracking is lost, and a new map is initialized in a fresh arbitrary scale. This is visible in our results as the near-total failure of the monocular configuration on all but the easy sequence, and as the ``fail'' entries in the control-point evaluation, where fewer than three control points could be triangulated. For the target domain, this method can be adapted by replacing the geometric front-end with learned detectors and matchers (e.g.\ SuperPoint and LightGlue), by adding CLAHE or deep-learning-based illumination enhancement upstream, and by tightly coupling an IMU so that the multi-map system can bridge the tracking gaps that currently fragment the trajectory.

DSO is limited by the brightness-constancy assumption at the heart of direct photometric alignment: it assumes that the intensity of a scene point is stable across consecutive frames. Low-light auto-exposure changes, sensor noise, and motion blur all violate this assumption, so the photometric residual no longer corresponds to true geometric misalignment. In our benchmark this manifests as large ATE on all but the easy sequence and outright failure on the hardest and one low-light sequence for RPE. The method could be partially adapted through a photometric calibration and online exposure compensation, and through robust or noise-aware photometric losses, but the underlying reliance on stable pixel intensities makes it fundamentally fragile in the operational conditions we target.

OpenVINS, as an MSCKF filter, is limited by linearization and the difficulty of maintaining a consistent covariance under strong non-linearities. When feature tracks are short-lived and outlier-ridden, as they are under handheld motion, occlusions, and moving platforms in the dataset, the filter diverges rather than merely drifting. This is the most severe behavior we observe: on the hard and one low-light sequence OpenVINS produces no usable initial interval at all, and its ATE exceeds that of the other methods well before the failure point. Adapting a filtering approach to this domain would require more robust outlier rejection, tighter feature-track quality gating, and illumination-robust features feeding the update step, though the filter's sensitivity to early estimation errors remains a structural weakness relative to optimization-based backends.

Kimera-VIO is our strongest contestant, but it is a visual-inertial \emph{odometry} system without loop closure in the configuration evaluated. Its limitation is therefore unbounded global drift: the RPE stays low across all sequences (local motion is well estimated thanks to the IMU), while the ATE grows with trajectory length because there is no mechanism to cancel accumulated error. This is exactly the pattern seen in Table~\ref{tab:lamaria-cp}, where Kimera-VIO scores high on the local CP@1m metric yet collapses on the global R@5m metric. The natural adaptation for the domain is to enable Kimera's full SLAM backend with loop closure and pose-graph optimization, so that revisited regions correct the accumulated drift that currently limits its global consistency.

DPVO and DPV-SLAM are the most robust systems in terms of tracking continuity, never losing tracking despite using neither inertial priors nor hand-crafted features. Their limitation is twofold. First, being monocular and inertial-free, they estimate trajectories only up to scale and lack the metric anchoring that the IMU gives Kimera; this contributes to their reduced accuracy relative to the successful VIO counterpart, visible in their large ATE on the low-light sequences. Second, as learning-based methods they are subject to a domain gap: their patch-tracking and update networks are trained predominantly on well-exposed imagery, so the severe noise, blur, and low contrast of our low-light sequences fall outside their training distribution, and the global backend of DPV-SLAM only helps when loop closures can actually be recognized under degraded appearance. These methods could be adapted by fine-tuning or augmenting training with low-light and blurred imagery, by prepending a learned enhancement front-end, and by fusing inertial measurements to resolve scale and provide a metric prior. The gap between DPVO and DPV-SLAM in Table~\ref{tab:lamaria-cp} also indicates that the loop-closure backend contributes little on these sequences, consistent with appearance-based loop detection struggling in the dark.

Finally, the evaluation itself carries limitations. We report only a subset of the LaMARia sequences for computational reasons, the low-light degradation in part of the data is synthetic and does not fully model camera exposure adaptation, and the monocular systems are only comparable to the metric ones after a $\mathrm{Sim}(3)$ alignment. These caveats should be kept in mind when generalizing the observed rankings to a deployed UGV platform.

\section{Conclusions}

This report surveys the background of modern SLAM systems, motivates the problem of low-light navigation for UGVs, and benchmarks six representative SLAM methods on the LaMARia dataset. 
We relate each method's performance to the assumptions underlying its paradigm, and survey algorithmic additions that could improve robustness in low-light conditions.

Future work includes implementing the proposed improvements within the evaluated SLAM approaches. 
We also plan to collect a dataset tailored specifically to the UGV domain and covering the problem of reduced illumination. 
This would enable a more focused evaluation of modern systems and may surface additional challenges in this domain that currently available datasets do not capture.

\section*{Acknowledgments}

This work was conducted as part of the certification program ``Off-Road Visual Navigation: Development and Evaluation of Systems in Challenging Environments'' at the Faculty of Applied Sciences, Ukrainian Catholic University, and in collaboration with the UCU UGV Club.

We express our gratitude to Maksym Zhuk, Anton Valihurskyi, and Andrii Kryvyi for their help and support throughout the project.

\bibliographystyle{plainnat}
\bibliography{ref}

\clearpage
\appendix
\section{SLAM Taxonomy}
\begin{figure}[ht]
    \centering
    \includegraphics[width=\columnwidth]{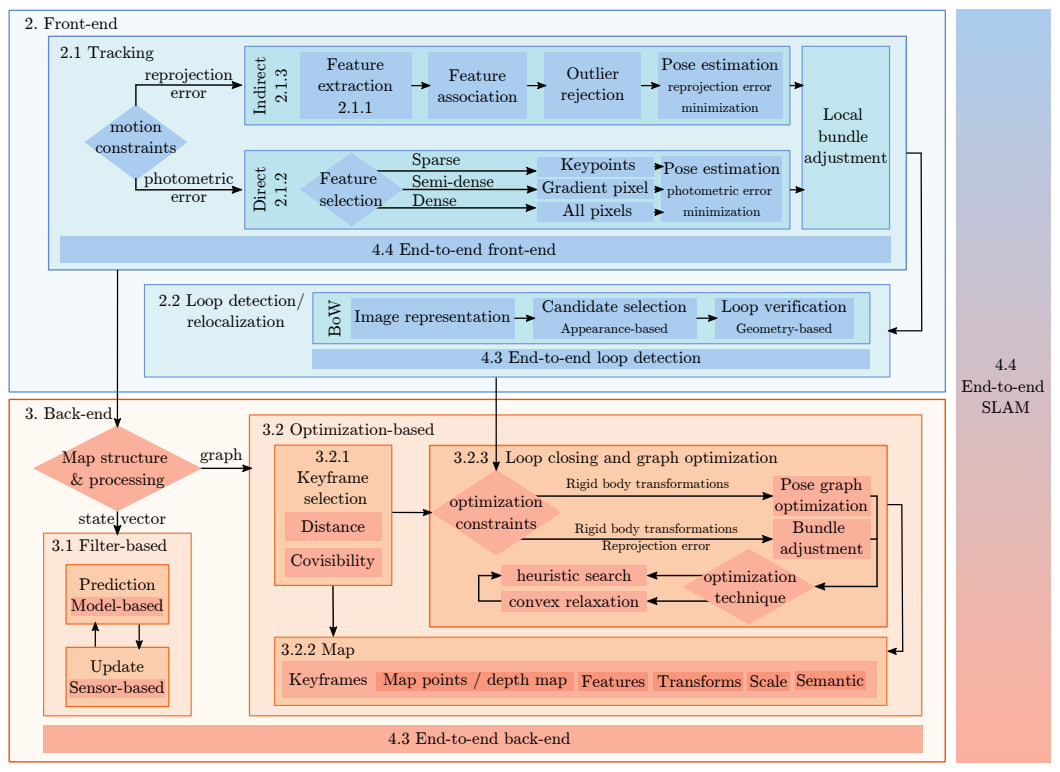}
    \caption{Standard modules for SLAM pipeline implementations as outlined in \citep{_lvarez_Tu_n_2024}}
\end{figure}

\section{Additional evaluation}

\subsection{Metrics}

LaMARia~\citep{krishnan2025benchmarkingegocentricvisualinertialslam} introduces additional metrics that evaluate consistency at surveyed ground-control points and across the broader pose sequence. 
These evaluation protocols use control points to assess both local accuracy at known locations and global consistency across the trajectory.

Each sequence is accompanied by $K$ geo-referenced control points $\{\mathbf{p}_k\}_{k=1}^{K} \subset \mathbb{R}^3$ whose positions are precisely determined through surveying techniques (e.g., GNSS-RTK measurements). 
These control points are automatically detected in imagery via fiducial markers. 
From the estimated trajectory and its keyframe observations, the estimated 3D position $\hat{\mathbf{p}}_k$ of each control point is triangulated by minimizing the reprojection error:

$$
E_{\text{tri}} = \sum_{i \in V(n)} \left\|\Pi\left(\mathbf{T}_i^{-1} \cdot \hat{\mathbf{p}}_k, C_i\right) - \mathbf{p}_i^n\right\|^2,
$$

where $\Pi(\cdot)$ is the camera projection function and $V(n)$ indexes the keyframes observing control point $n$. 
Then this is solved for an optimal $\text{Sim}(3)$ transformation $S^\star$ that aligns the estimated control points with the control points surveyed by robust least-squares optimization, accounting for unknown scale, rotation, and translation.

The fraction of control points whose aligned estimates fall within $\tau$ metres of their surveyed positions:

$$\mathrm{CP}@\tau = \frac{1}{K} \sum_{k=1}^{K} \mathbb{1}\left[\left\|S^\star\hat{\mathbf{p}}_k - \mathbf{p}_k\right\| \leq \tau\right].$$
We report $\mathrm{CP}@1\text{m}$. Because $S^\star$ is estimated from the control points themselves, this metric measures local accuracy at surveyed locations and is insensitive to trajectory drift in unsurveyed regions. 

As suggested in LaMARia~\citep{krishnan2025benchmarkingegocentricvisualinertialslam}, we compute dense pseudo-ground-truth poses by fusing visual, inertial, and control-point information through joint optimization to evaluate global trajectory consistency. 
Let $\{\mathbf{T}_j^{\mathrm{pGT}}\}_{j=1}^{L}$ denote the set of $L$ pseudo-ground-truth poses matched (within temporal tolerance) to estimated keyframes.

Using the same alignment $S^\star$ computed from control points, we measure the fraction of poses whose estimated locations align well with pseudo-ground-truth:

$$
\mathrm{R}@\tau = \frac{1}{L} \sum_{j=1}^{L} \mathbb{1}\left[\left\|\mathrm{trans}\left(\left(S^\star\hat{\mathbf{T}}_j\right)^{-1} \mathbf{T}_j^{\mathrm{pGT}}\right)\right\| \leq \tau\right],
$$
where $\mathrm{trans}(\mathbf{T})$ extracts the translational component of a transformation. We report $\mathrm{R}@5\text{m}$.

These two metrics reveal different failure modes of the systems. 
A trajectory showing high $\mathrm{CP}@1\text{m}$ but low $\mathrm{R}@5\text{m}$ indicates local consistency at surveyed control points with global drift elsewhere.
Conversely, uniform scale or orientation errors affect both metrics equally. 

\subsection{Results}

Table~\ref{tab:lamaria-cp} reports $\mathrm{CP}@1\text{m}$ and $\mathrm{R}@5\text{m}$ for the two sequences that carry surveyed control points. Table~\ref{tab:lamaria-cp-sens} breaks down the $\mathrm{R}@5\text{m}$ denominator/tolerance choice for the systems that produced a usable estimate. The strict 1\,ms \emph{full} column is deflated by time-grid mismatch alone (Kimera-VIO and DPVO drop to roughly 50--80\% coverage), while the looser $\sim$100\,ms \emph{cov} tolerance restores coverage to near 100\% with $\mathrm{cov}\approx\mathrm{assoc}$, confirming these systems track the full sequence and that \emph{assoc} is the fair comparison figure; OpenVINS is unaffected, as its poses already fully overlap the pGT.

\begin{table}[t]
\centering
\small\setlength{\tabcolsep}{4pt}
\begin{tabular}{l ccc ccc}
\toprule
& \multicolumn{3}{c}{seq.\ 4\_10} & \multicolumn{3}{c}{seq.\ 4\_11} \\
\cmidrule(lr){2-4}\cmidrule(lr){5-7}
Method & score & CP@1m & R@5m & score & CP@1m & R@5m \\
\midrule
Kimera-VIO & 100.0 & 100.0 & 0.0 & 94.2 & 100.0 & 2.5 \\
OpenVINS & 80.8 & 81.2 & 0.0 & 31.3 & 13.3 & 41.3 \\
DSO & \multicolumn{3}{c}{fail} & \multicolumn{3}{c}{fail} \\
ORB-SLAM3 & \multicolumn{3}{c}{fail} & \multicolumn{3}{c}{fail} \\
DPVO & 72.2 & 75.0 & 5.6 & 36.7 & 33.3 & 6.6 \\
DPV-SLAM & 34.8 & 37.5 & 0.2 & 32.1 & 26.7 & 0.2 \\
\midrule
\emph{pGT (ceiling)} & 82.4 & 100.0 & 100.0 & 91.0 & 100.0 & 100.0 \\
\bottomrule
\end{tabular}
\caption{Control-point and pseudo-GT recall on the two LaMARia sequences with survey control points, after Sim(3) alignment solved from the control points. CP@1m is local accuracy at the control points; R@5m is global pose recall. ``fail'' = fewer than 3 control points triangulated. pGT row is the achievable ceiling.}
\label{tab:lamaria-cp}
\end{table}

\begin{table}[t]
\centering
\small\setlength{\tabcolsep}{3.5pt}
\begin{tabular}{l cccc cccc}
\toprule
& \multicolumn{4}{c}{seq.\ 4\_10} & \multicolumn{4}{c}{seq.\ 4\_11} \\
\cmidrule(lr){2-5}\cmidrule(lr){6-9}
Method & assoc & full & cov & cov\% & assoc & full & cov & cov\% \\
\midrule
Kimera-VIO & 0.0 & 0.0 & 0.0 & 99 & 2.5 & 0.6 & 3.1 & 100 \\
OpenVINS & 0.0 & 0.0 & 0.0 & 100 & 41.3 & 41.2 & 41.2 & 100 \\
DSO & \multicolumn{4}{c}{fail} & \multicolumn{4}{c}{fail} \\
ORB-SLAM3 & \multicolumn{4}{c}{fail} & \multicolumn{4}{c}{fail} \\
DPVO & 5.6 & 2.8 & 5.5 & 100 & 6.6 & 3.2 & 6.5 & 100 \\
DPV-SLAM & 0.2 & 0.1 & 0.2 & 100 & 0.2 & 0.1 & 0.2 & 100 \\
\midrule
\emph{pGT (ceiling)} & 100.0 & 100.0 & 100.0 & 100 & 100.0 & 100.0 & 100.0 & 100 \\
\bottomrule
\end{tabular}
\caption{R@5m sensitivity to the matching tolerance. \emph{assoc}: associated pGT poses at 1\,ms. \emph{full}: full pGT length at 1\,ms. \emph{cov}: full pGT length at $\sim$100\,ms tolerance; cov\% is resulting coverage. See text for interpretation.}
\label{tab:lamaria-cp-sens}
\end{table}

\section{Trajectories}

We visualize the tracked trajectories against ground truth for the \verb|sequence_4_10|, \verb|sequence_4_11|, \verb|R_06_medium|, and \verb|R_09_hard| sequences. OpenVINS demonstrates a catastrophic failure mode on the \verb|R_09_hard| and \verb|sequence_4_10| datasets, with trajectories that abruptly veer off-map mid-sequence. The method tracks correctly for the first 150--200 meters but then crosses a discrete scale observability threshold at a specific spatial location, after which all subsequent poses are computed in a warped coordinate frame with no recovery mechanism.

Direct Sparse Odometry exhibits geometric impossibility on sequences \verb|sequence_4_10| and \verb|sequence_4_11|. Early depth estimation propagates scale bias throughout the sequence, causing scale to oscillate between over- and under-estimation, which forces camera pose estimates to bounce back and forth, creating topological destruction.

OpenVINS on \verb|sequence_4_11| weaves perpendicular to the actual path while maintaining good local tracking but accumulating large global error. The root cause is anisotropic scale bias: the method estimates forward motion (aligned with dominant camera motion) more accurately than lateral motion, creating systematic weaving perpendicular to travel direction.

Kimera-VIO maintains topological correctness on all sequences, staying on-path with only minor jitter and no spirals, with gradual ATE growth. Overall, the method shows a graceful degradation rather than catastrophic failure.

Scale observability is the limiting factor for all methods on longer sequences. Once visual odometry incorrectly estimates scale, no mechanism exists to correct it without loop closure or IMU integration, making scale the fundamental constraint determining method performance.

\begin{figure}[ht]
    \centering
    \includegraphics[width=\columnwidth]{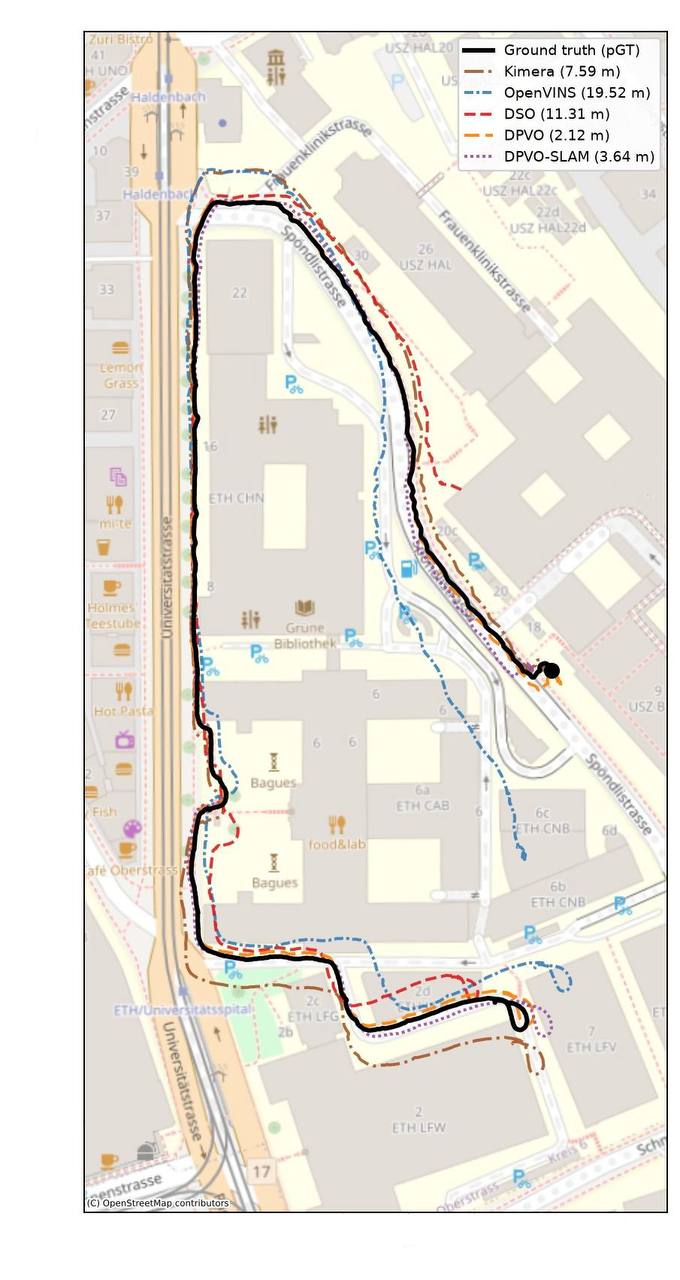}
    \caption{Estimated versus ground-truth trajectories on \texttt{R\_06\_medium}. Accumulated drift grows gradually with trajectory length for the odometry-only methods, while Kimera-VIO stays on-path, consistent with the ATE trend in Table~\ref{tab:method_comparison}.}
    \label{fig:traj-medium}
\end{figure}

\begin{figure}[ht]
    \centering
    \includegraphics[width=\columnwidth]{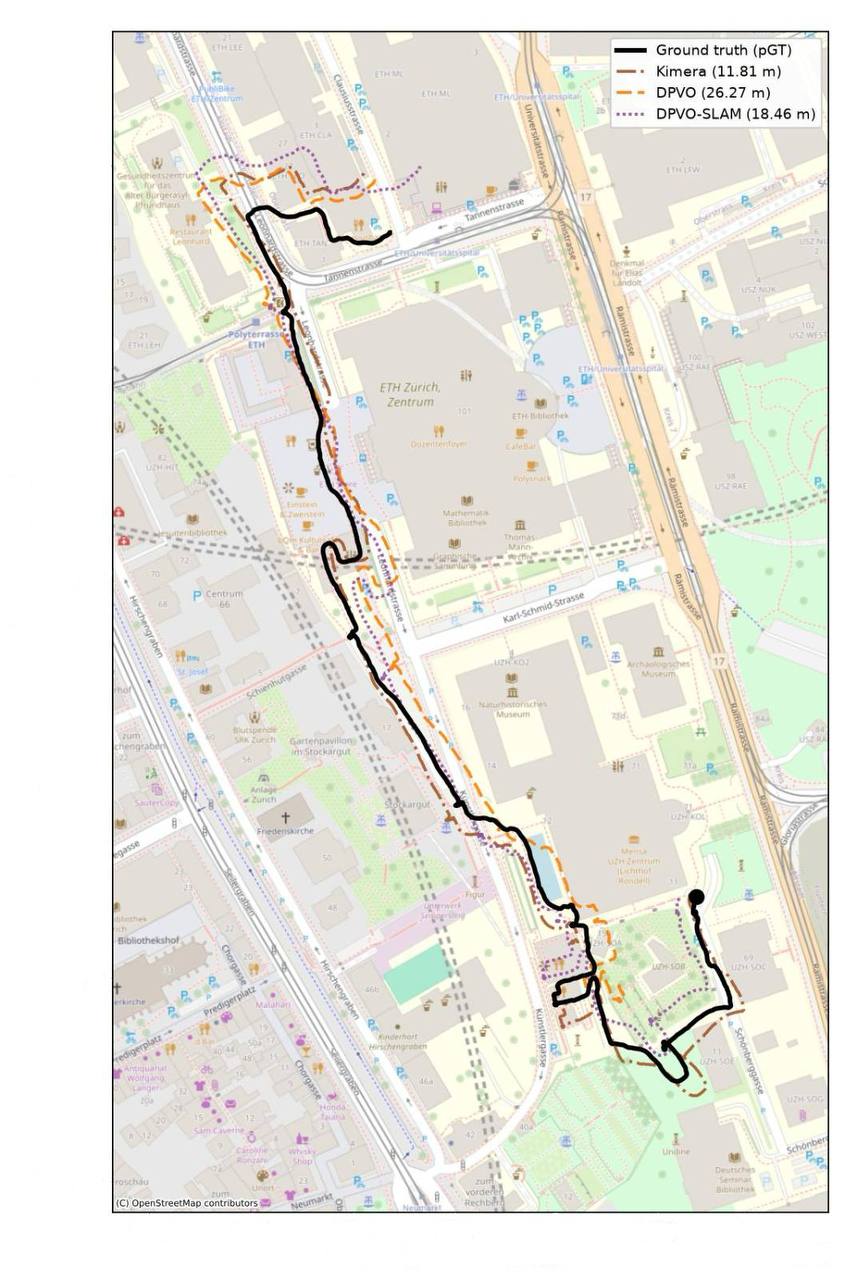}
    \caption{Estimated versus ground-truth trajectories on \texttt{R\_09\_hard}. OpenVINS veers off-map after crossing a scale-observability threshold, and the feature-based and direct methods lose tracking, whereas DPVO, DPV-SLAM, and Kimera-VIO follow the full path with varying global drift.}
    \label{fig:traj-hard}
\end{figure}

\begin{figure}[ht]
    \centering
    \includegraphics[width=\columnwidth]{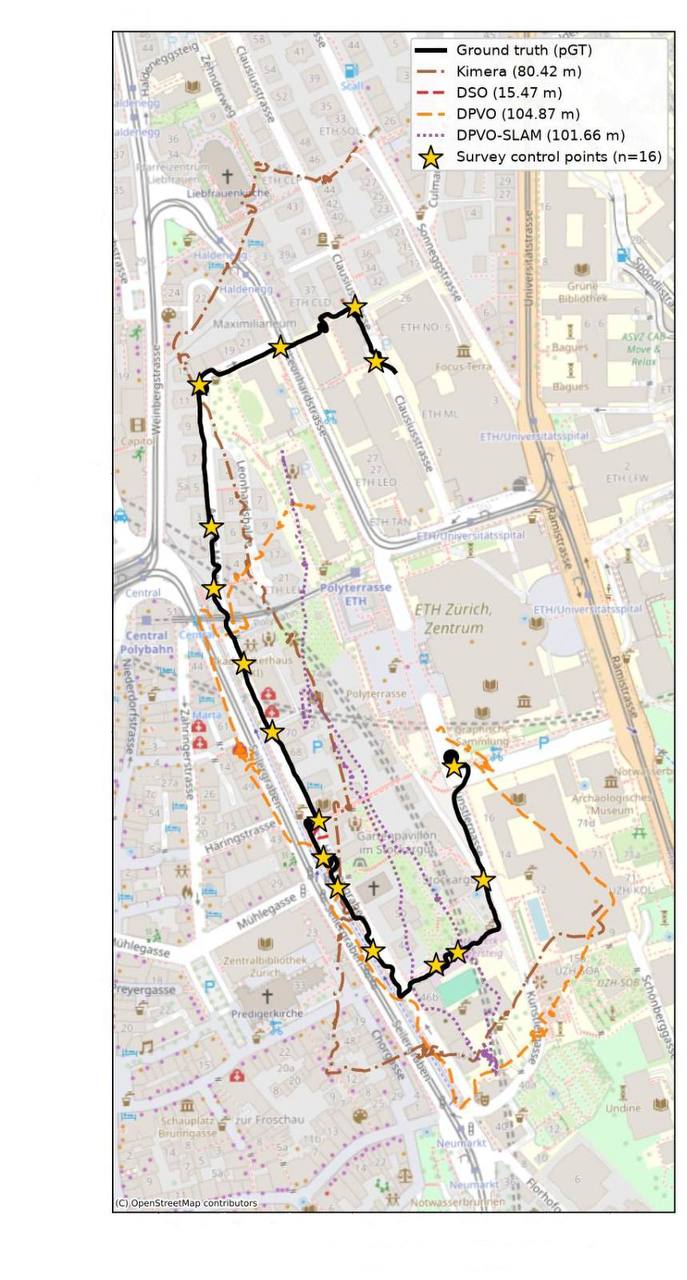}
    \caption{Estimated versus ground-truth trajectories on the low-light sequence \texttt{sequence\_4\_10}. Under severe illumination degradation OpenVINS diverges into a warped frame and DSO exhibits oscillating scale, while the monocular learning-based methods maintain tracking at the cost of large scale and drift error.}
    \label{fig:traj-ll1}
\end{figure}

\begin{figure}[ht]
    \centering
    \includegraphics[width=\columnwidth]{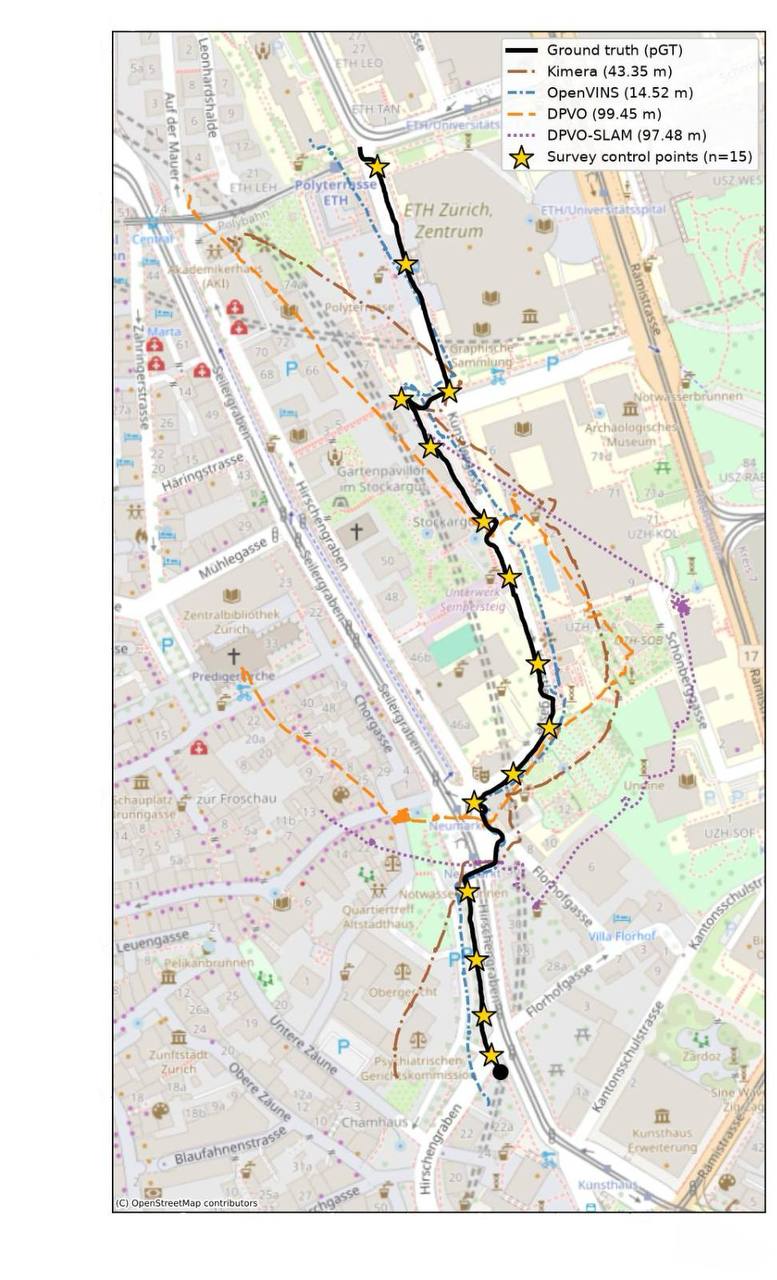}
    \caption{Estimated versus ground-truth trajectories on the low-light sequence \texttt{sequence\_4\_11}. OpenVINS weaves perpendicular to the path due to anisotropic scale bias, illustrating the divergence between local motion accuracy (RPE) and global consistency (ATE) discussed in the results.}
    \label{fig:traj-ll2}
\end{figure}

\end{document}